\documentclass[10pt,a4paper]{article}
\pdfoutput=1
\usepackage{lrec-coling2024}\usepackage{amsmath}\usepackage{graphicx}\usepackage{subfigure}\title{Principal Component Analysis as a Sanity Check\\for Bayesian Phylolinguistic Reconstruction}\name{Yugo Murawaki}\address{Graduate School of Informatics, Kyoto University\\Yoshida-honmachi, Sakyo-ku, Kyoto, 606-8501, Japan\\murawaki@i.kyoto-u.ac.jp\\}\abstract{Bayesian approaches to reconstructing the evolutionary history of languages rely on the tree model, which assumes that these languages descended from a common ancestor and underwent modifications over time. However, this assumption can be violated to different extents due to contact and other factors. Understanding the degree to which this assumption is violated is crucial for validating the accuracy of phylolinguistic inference. In this paper, we propose a simple sanity check: projecting a reconstructed tree onto a space generated by principal component analysis. By using both synthetic and real data, we demonstrate that our method effectively visualizes anomalies, particularly in the form of \textit{jogging}.\\\newline}\begin{document}\maketitleabstract\section{Introduction}The tree model serves as the foundation for historical-comparative linguistics~\citep{a}. Although manual inference has traditionally been dominant in the field~\cite{b}, the influence of evolutionary biology has led to a rapid rise to computation-heavy statistical analysis of linguistic data~\citep{c,d,e,f}, spawning a multitude of papers built upon Bayesian phylolinguistic tools.

The tree model assumes that the evolutionary history of related languages can be represented as a tree. The root represents a single common ancestor and a number of branching events lead to the observed languages. Over time, modifications gradually accumulate along the branches, indicating that the distance between two languages on the tree approximately corresponds to the extent of divergence between them. Various methods have been proposed based on this intuition to address the inverse problem of reconstructing the tree from the observed languages~\citep{g}.

In reality, the tree model is violated to varying degrees. When languages come into contact, there is often a horizontal transmission of features between them, despite the assumption that they evolve independently. This horizontal transmission necessitates the addition of extra edges, resulting in a representation that is no longer a tree but a network~\citep{h,i}. Despite efforts to integrate horizontal transfer into statistical models~\cite{j,k}, achieving stable and scalable inference continues to pose a significant challenge. For this reason, the tree model retains its dominant position in phylolinguistics.

The modern proponents of the tree model are well aware of repeated criticisms that in fact date back centuries \cite{l,m}. Initially, they attempted to demonstrate the model's robustness against horizontal transmission by utilizing synthetic data~\citep{n,o}. Subsequently, they focused their attention on examining the extent to which the tree model is applicable to real data~\citep{p,q}. Unfortunately, there is a disparity between the tree model and their analytical tools: NeighborNet~\citep{r}, the $\delta$ score~\citep{s}, and the $Q$-residual score~\citep{p}. These tools are all based on distance-based approaches despite the use of Bayesian methods for phylolinguistic reconstruction.\begin{figure*}[t]\centering\includegraphics[width=1.0\textwidth]{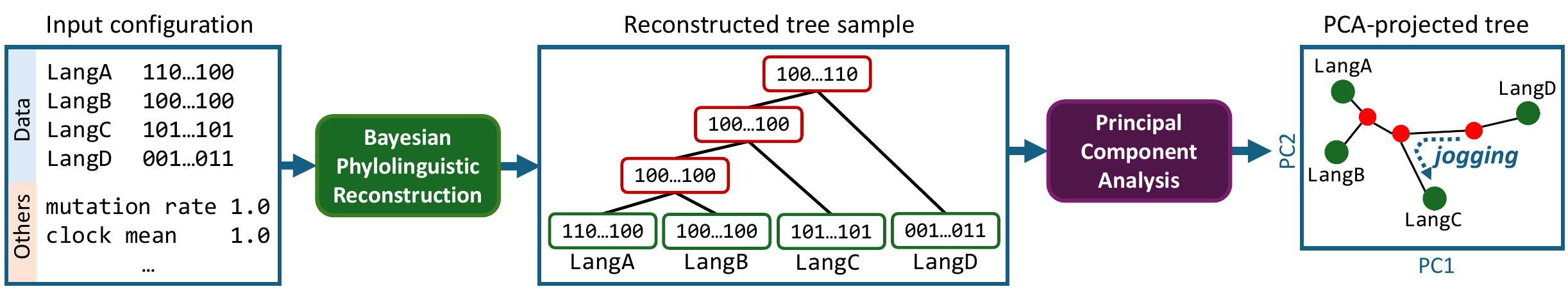}\caption{Overview of the proposed method. In this example, we reconstruct a phylogenetic tree for four modern languages, resulting in three ancestral nodes with explicitly represented states. The states of these seven languages are then subjected to principal component analysis (PCA), followed by projection onto a low-dimensional space. The downward path from the root to \texttt{LangC} exhibits \textit{jogging}.}\label{fig:a}\end{figure*}

Recent studies~\citep{q} conduct additional analyses using Bayesian tree summarization tools such as DensiTree~\citeplanguageresource{t}, based on the speculation that a relative absence of disagreements within a summary tree may indicate endorsement of the tree model. However, uncertainty is an intrinsic characteristic of Bayesian inference that emerges regardless of whether the model's assumptions are valid. After all, the model itself lacks a direct means to assess the accuracy of its underlying assumptions. It indeed can be deceived by fundamentally non-tree-like generative processes~\citep{u}.

In this paper, we present a simple and practical approach to directly analyzing Bayesian phylolinguistic reconstruction. We apply principal component analysis (PCA) to language states and project a reconstructed tree onto the PCA-generated space (Figure~\ref{fig:a}). Our key idea is to leverage \textit{continual diversification}, an aspect of tree-shaped evolution that usually falls outside the scope of the model's assumptions. We expect ancestor-descendant transitions to follow a unidirectional pattern along the first principal component axis. A gross violation of this unidirectionality, which we call \textit{jogging}, can be seen as a deviation from the tree model, as is evident in Figure~\ref{fig:e}. To illustrate the usefulness of the proposed method, we provide demonstrations using both synthetic and real data, emphasizing its potential as a sanity check. The code is publicly available at \url{https://github.com/murawaki/treepca}.\section{Preliminaries}\subsection{Binary Sequence Representations}In a typical Bayesian phylolinguistic reconstruction scenario, we are provided with a collection of observed languages, where each language is represented as a binary sequence. Most studies use binary-coded basic vocabulary data. These lexical data are originated from glottochronology~\citep{v}, despite the decline of glottochronology itself due to substantial criticism~\citep{w}.

Basic vocabulary items such as \texttt{WATER} and \texttt{EAT} are assumed to be culture independent and resistant to change. The process of constructing lexical data involves two steps. Linguists begin by collecting words for these items in each language. Subsequently, they assess the cognacy (relatedness) of these words across languages. For example, English \textit{water} and German \textit{Wasser} share their etymological root whereas French \textit{eau} and Italian \textit{acqua} are cognates. By organizing these two cognate groups, we can represent English and German as \texttt{10} and French and Italian as \texttt{01}, where \texttt{1} and \texttt{0} indicate the presence and absence of a cognate group, respectively. Concatenating multiple basic vocabulary items, we typically obtain hundreds or thousands of binary features.

Although the evolutionary process of these binary features is assumed to follow a tree-like pattern, this assumption is not exempt from violations. One common type of deviation arises from loanwords. Since cognacy judgments rely on regular sound correspondences, loanwords can be identified by linguists and subsequently excluded from the dataset. However, older loanwords and borrowings between closely-related languages pose a higher risk of going undetected, thus potentially eluding removal from the analysis.

Thanks to the arbitrariness of meaning-symbol connection, it is generally assumed that a feature is gained only once throughout history. However, it is important to recognize that this assumption can be violated. One common cause of such deviations is semantic shift. For instance, the semantic shift from \texttt{PERSON} to \texttt{MAN} is universal and can happen in parallel, leading to multiple gains of the same word for \texttt{MAN} in a tree~\citep{x}.\subsection{Bayesian Phylolinguistic Models}\label{sec:b}Bayesian phylolinguistic models encompass a range of advanced statistical techniques. For a comprehensive understanding of the details, we recommend referring to \citet{y}. Here, we will provide a high-level overview of the topic.

A phylolinguistic model assigns a probability to a generative process that begins with a common ancestor and extends to observed languages.\footnote{To be precise, coalescent variants of the time-tree model look backward in time.} The probabilistic assessment can be subdivided into three primary components: a time-tree, state transitions, and rate variations. A time-tree represents a rooted tree where each node is associated with a calendar or relative date. The likelihood of a given time-tree is evaluated using a time-tree model.

Each node holds a binary sequence as its state, and the transition from a parent to a child involves gains (\texttt{0} $\rightarrow$ \texttt{1}) and losses (\texttt{1} $\rightarrow$ \texttt{0}). The probability of such transitions is assessed by a continuous-time state transition model. For inference efficiency, the states of the unobserved languages are usually marginalized out, accounting for all possible combinations of the states~\citep{z}.

The state transition model is linked to a rate model. The strict clock model enforces a uniform rate of change in a tree, whereas various relaxed clock models investigate rate variations. By assigning different rates to different branches, we can analyze the potential alternation of rapid and slow phases of language change~\citep{aa}. Furthermore, assigning distinct rates to features or groups of features allows for the exploration of the hypothesis that certain vocabulary items display greater stability~\citep{ab}.

With observed languages and optional hard constraints, the remaining portion of the generative process defines the search space. The conventional inference method is Markov Chain Monte Carlo (MCMC) sampling, which generates samples from the probability distribution. For our analysis, it is important to note that the sampler does not directly track the states of the unobserved languages because they are marginalized out. Nevertheless, it is easy to generate them using an algorithm analogous to forward filtering-backward sampling for sequence data~\citep{ac}.

MCMC sampling yields a vast number of time-trees, making it necessary to employ summarization techniques for human interpretation. One widely used approach is to construct a maximum clade credibility (MCC) tree by merging these samples. DensiTree~\citeplanguageresource{t} offers another type of intuitive visualization that effectively highlights disagreements among the samples.\subsection{Principal Component Analysis (PCA)}Principal component analysis (PCA) linearly transforms high-dimensional data into a new coordinate system, where each principal component (PC) represents a new axis. Since the first few PCs usually capture key variance in the original data, PCA can be used for visualization.

While usually deemed irrelevant in phylolinguistics, PCA is ubiquitous in population genetics~\citep{ad,ae}. PCA itself is agnostic to the evolutionary process underlying genome data. In fact, whole-genome data do not follow a tree-like pattern either at a micro level due to recombination or at a macro level due to admixture (interbreeding of distinct populations). While population genetics gives weight to scalability~\citep{af}, a na\"{i}ve implementation suffices for small linguistic data.

Formally, let $\widetilde{X}$ be an $n \times p$ binary matrix, where $n$ is the number of languages and $p$ is the number of features. We first apply mean centering to $\widetilde{X}$:\[X = \widetilde{X} - \boldsymbol{\mu},\]where each element $\mu_i$ of the vector $\boldsymbol{\mu}$ represents the mean of the corresponding feature. We then apply singular value decomposition (SVD) to $X$:\[X = U \Sigma V^\mathsf{T},\]where $U$ is an $n \times n$ orthogonal matrix containing the left singular vectors, $\Sigma$ is an $n \times p$ diagonal matrix of singular values, and $V^\mathsf{T}$ is the transpose of an $p \times p$ orthogonal matrix containing the right singular vectors. Finally, we obtain the projection of $X$ onto the $i$-th PC by \[\hat{\boldsymbol{x}}_{i} = X \boldsymbol{u}_i,\]where $\boldsymbol{u}_i$ is the $i$-th column of $U$.\footnote{Contrary to a belief mentioned in \citet{ag}, PCA does not preserve distances between data points in the lower-dimensional space. For cases where distance preservation is crucial, which we suspect might not be prevalent, considering multidimensional scaling (MDS) may be more appropriate.}

The proportion of variance explained by the $i$-th PC can be calculated as $\lambda_i / \sum_{i=1}^{k} \lambda_i$, where $\lambda_i = \sigma_i^2 / (n-1)$. In this paper, we only use the first two PCs for visualization. In fact, the proportion of variance explained by the first two PCs for linguistic data (usually a few tens of percent) is much larger than that for genome data.\section{Proposed Method}Our idea is fairy simple: project a reconstructed tree onto the two-dimensional space generated by PCA to check if it exhibits anomalies. To do this, we begin by applying PCA to $\widetilde{X}$, the states of the observed languages, to calculate $\hat{\boldsymbol{x}}_{1}$, $\hat{\boldsymbol{x}}_{2}$, $\boldsymbol{\mu}$, $\boldsymbol{u}_1$, and $\boldsymbol{u}_2$.\footnote{Observed languages may contain missing features. In that case, we let the sampler impute these values.} Next, We perform Bayesian phylolinguistic reconstruction and obtain a sample tree from the sampler. Let $\widetilde{Y}$ be an $(n-1) \times p$ matrix representing the states of the unobserved languages in the sample.\footnote{A bifurcating tree with $n$ leaves has $n-1$ internal nodes including the root.} Using $\boldsymbol{\mu}$, $\boldsymbol{u}_1$, and $\boldsymbol{u}_2$, we map $\widetilde{Y}$ to $\hat{\boldsymbol{y}}_{1}$ and $\hat{\boldsymbol{y}}_{2}$. Finally, we draw a scatter plot of the entire set of languages, with additional straight lines connecting parents to children.\begin{figure*}[t]\centering\subfigure[No borrowing.]{\includegraphics[width=0.48\textwidth]{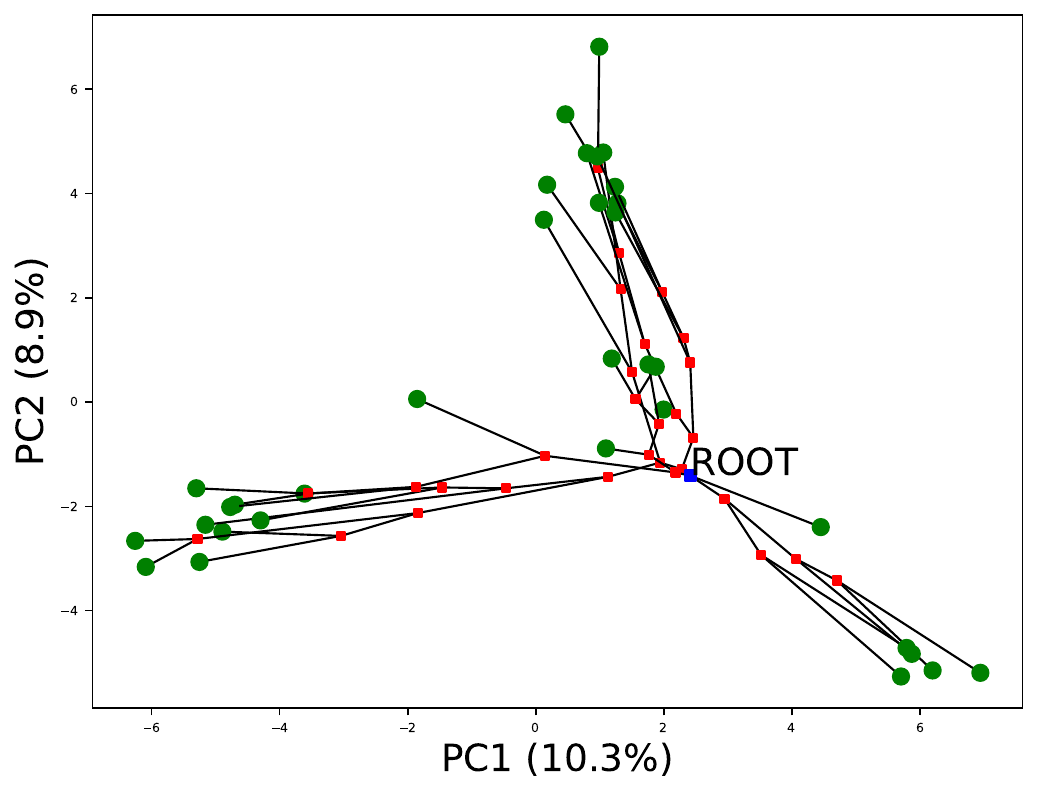}}\hfill\subfigure[Global borrowing.]{\includegraphics[width=0.48\textwidth]{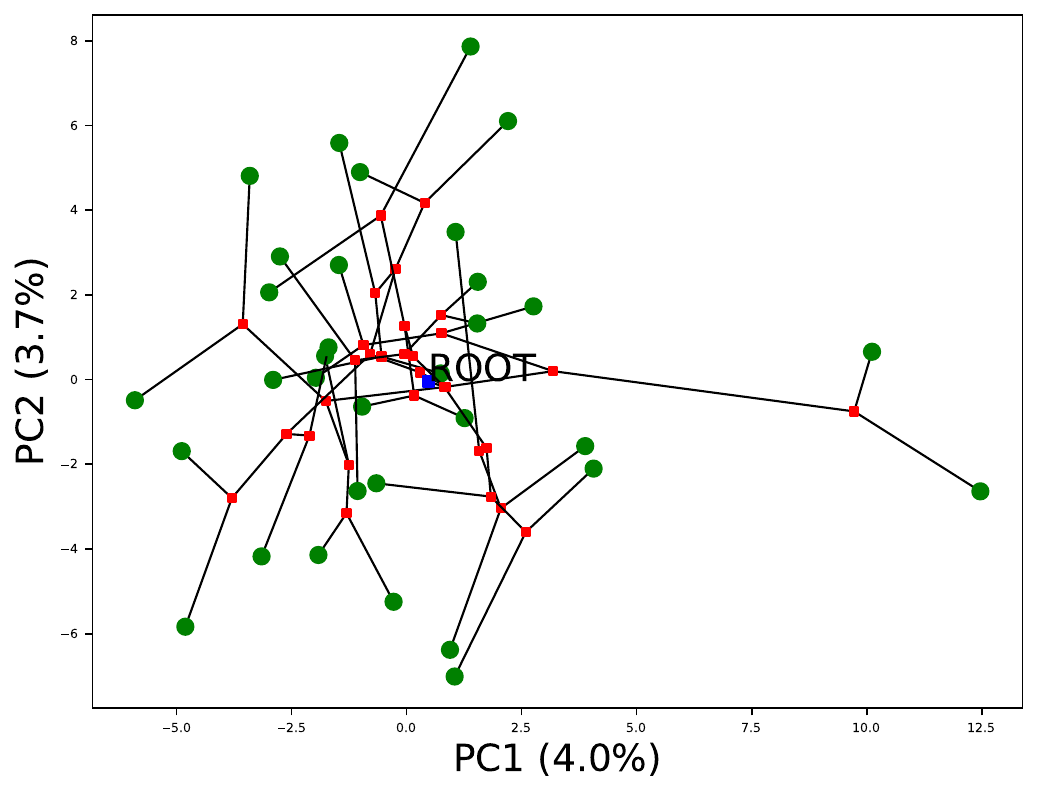}}\\\subfigure[Local borrowing (1,000-year limit).]{\includegraphics[width=0.48\textwidth]{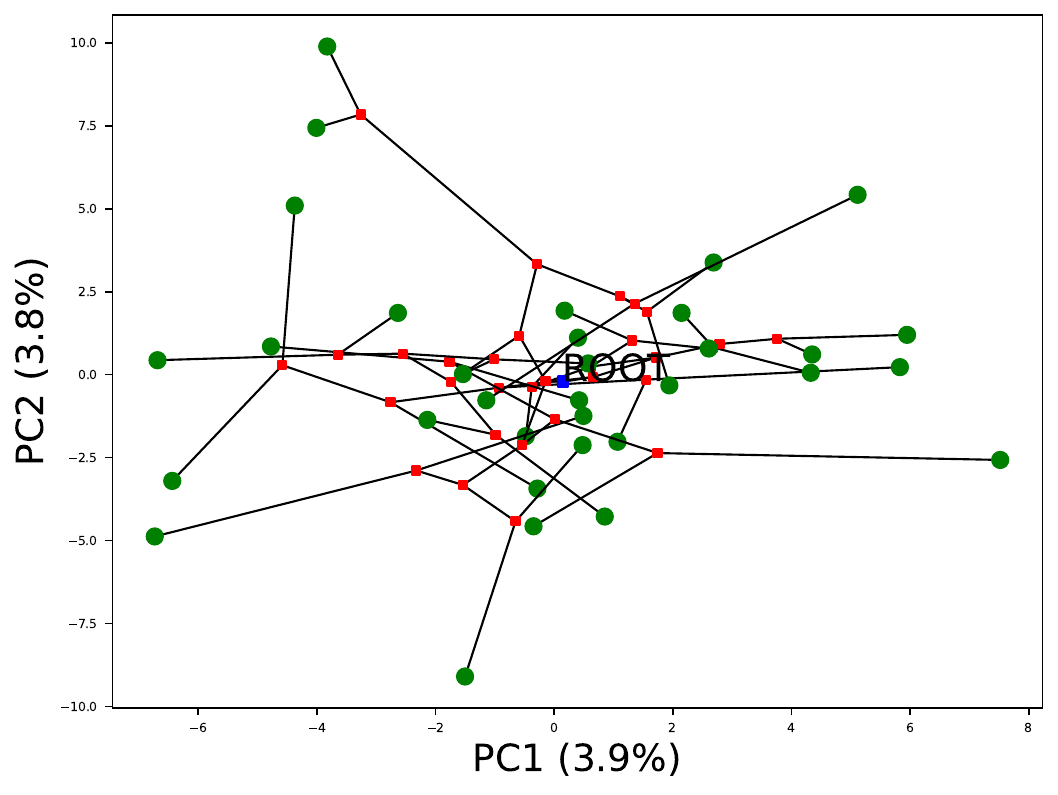}}\hfill\subfigure[Local borrowing (3,000-year limit).]{\includegraphics[width=0.48\textwidth]{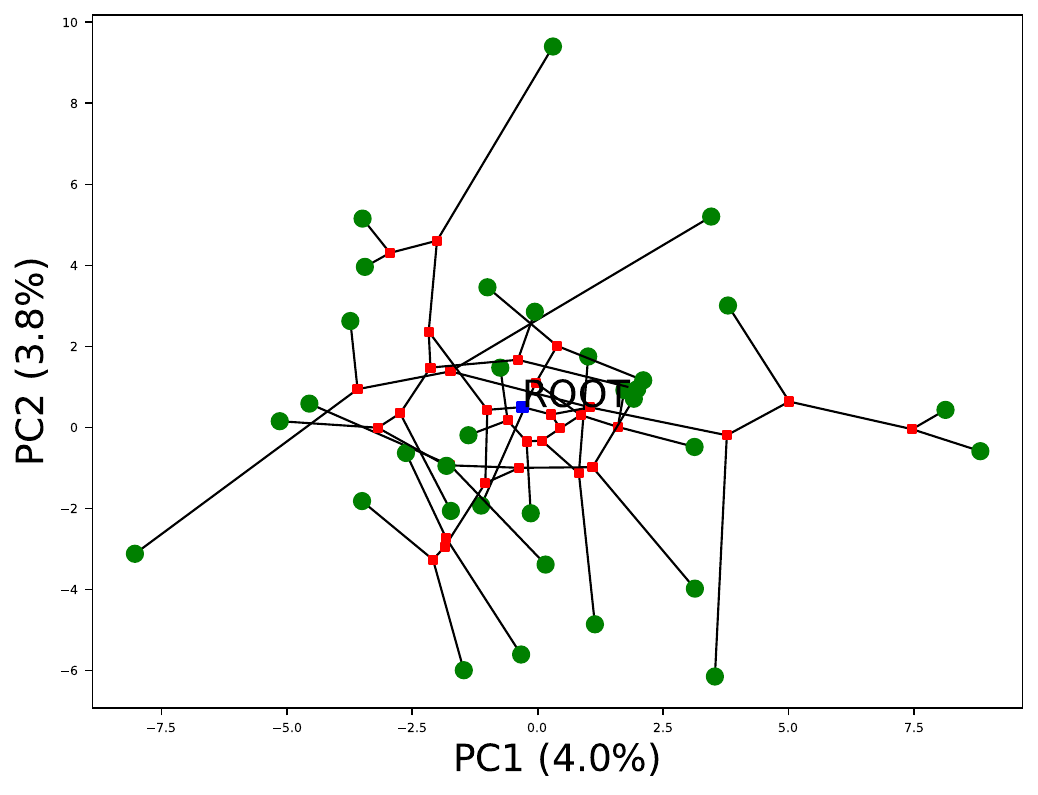}}\caption{PCA of Bayesian phylolinguistic reconstruction for the skewed time-tree of data simulation, with four borrowing scenarios. We used the first two PCs, denoted as PC1 and PC2. A percentage indicates the amount of variance explained by the corresponding PC. Circles indicate observed leaf nodes while rectangles denote reconstructed internal nodes.}\label{fig:c}\end{figure*}\begin{figure*}[t]\centering\subfigure[No borrowing.]{\includegraphics[width=0.48\textwidth]{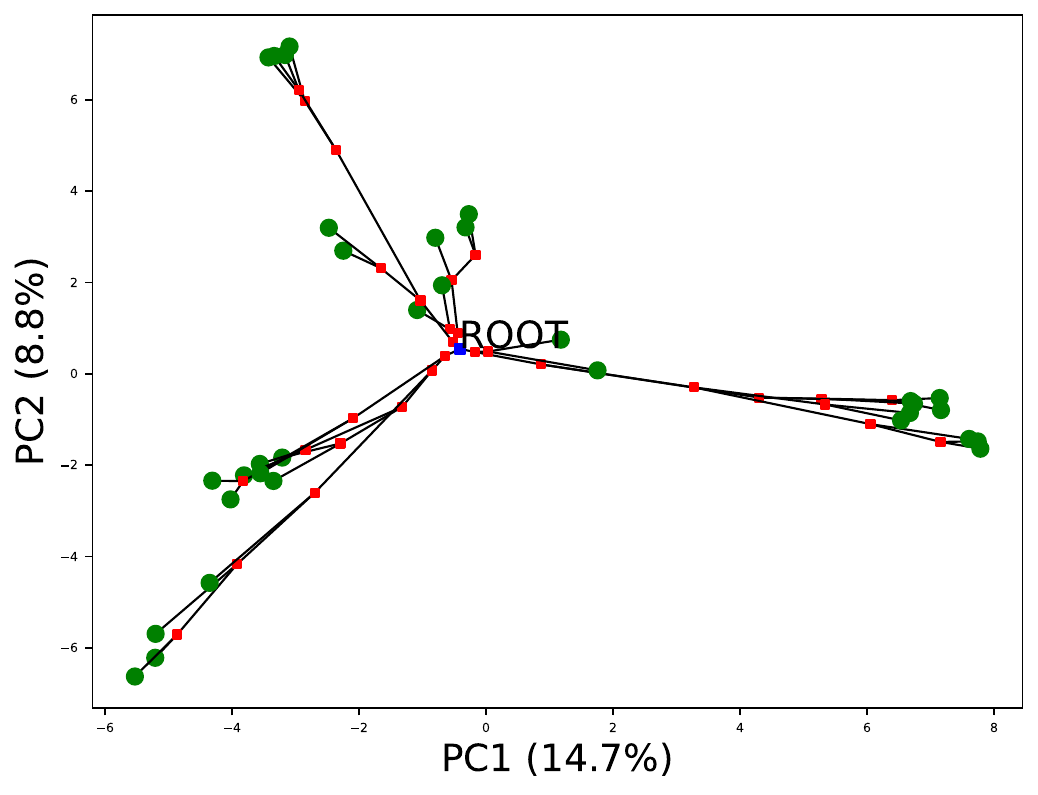}}\hfill\subfigure[Global borrowing.]{\includegraphics[width=0.48\textwidth]{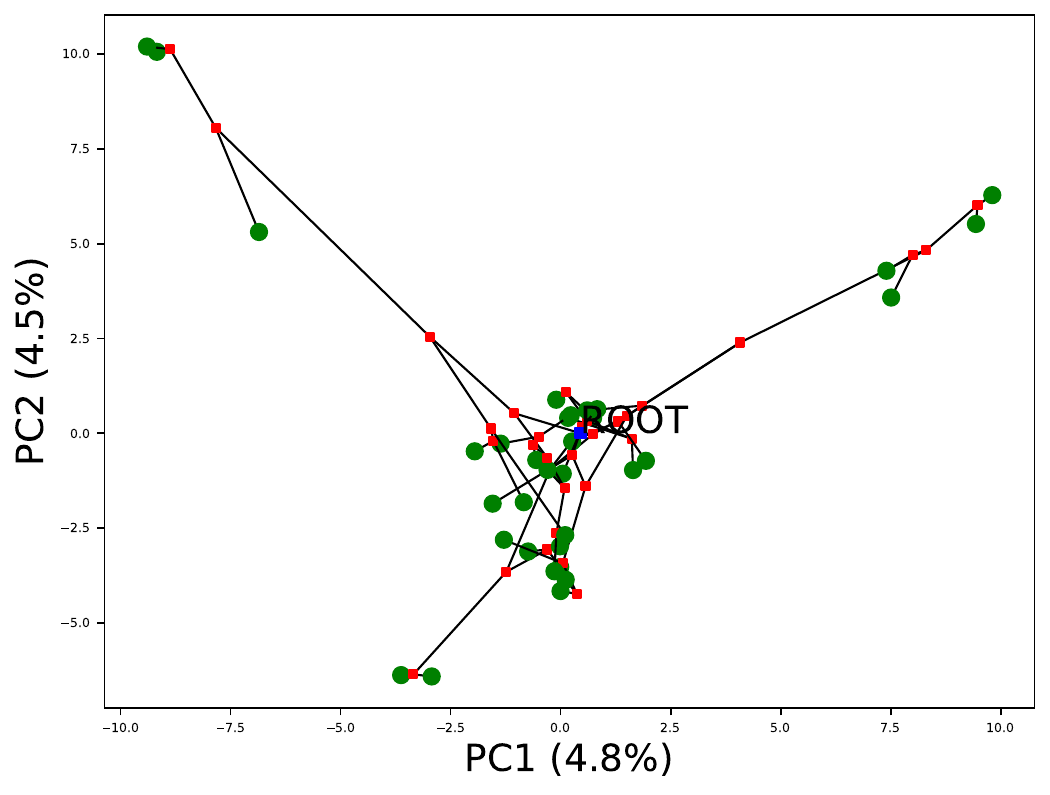}}\\\subfigure[Local borrowing (1,000-year limit).]{\includegraphics[width=0.48\textwidth]{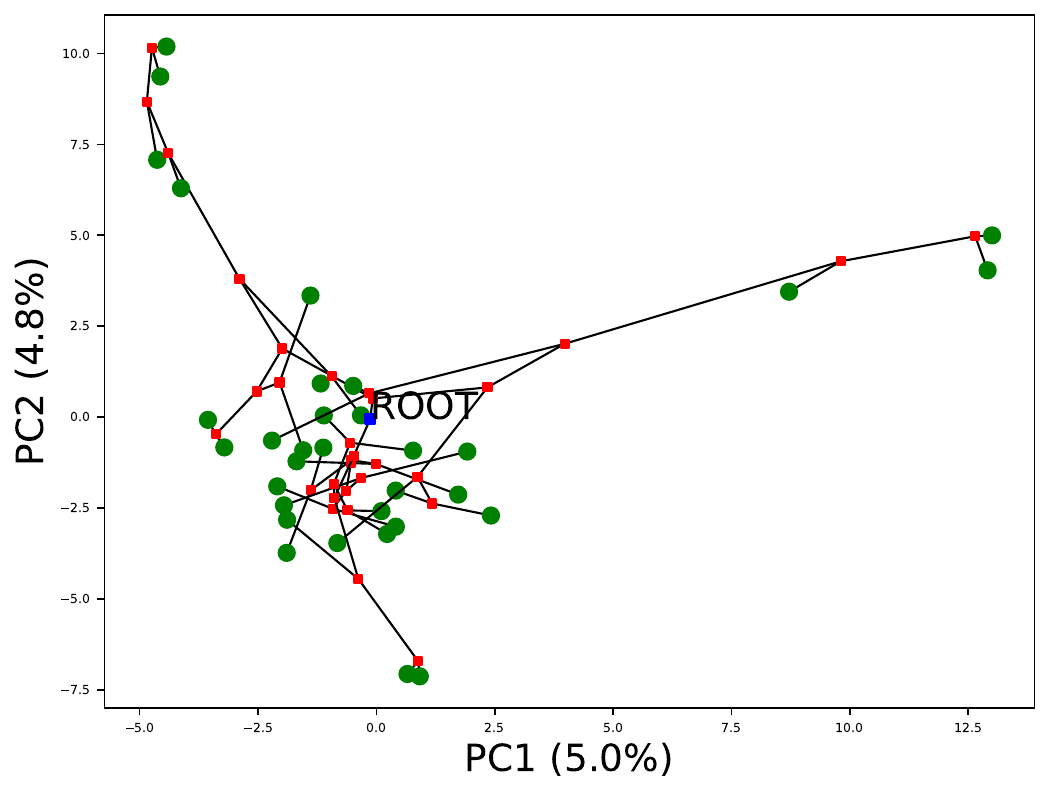}}\hfill\subfigure[Local borrowing (3,000-year limit).]{\includegraphics[width=0.48\textwidth]{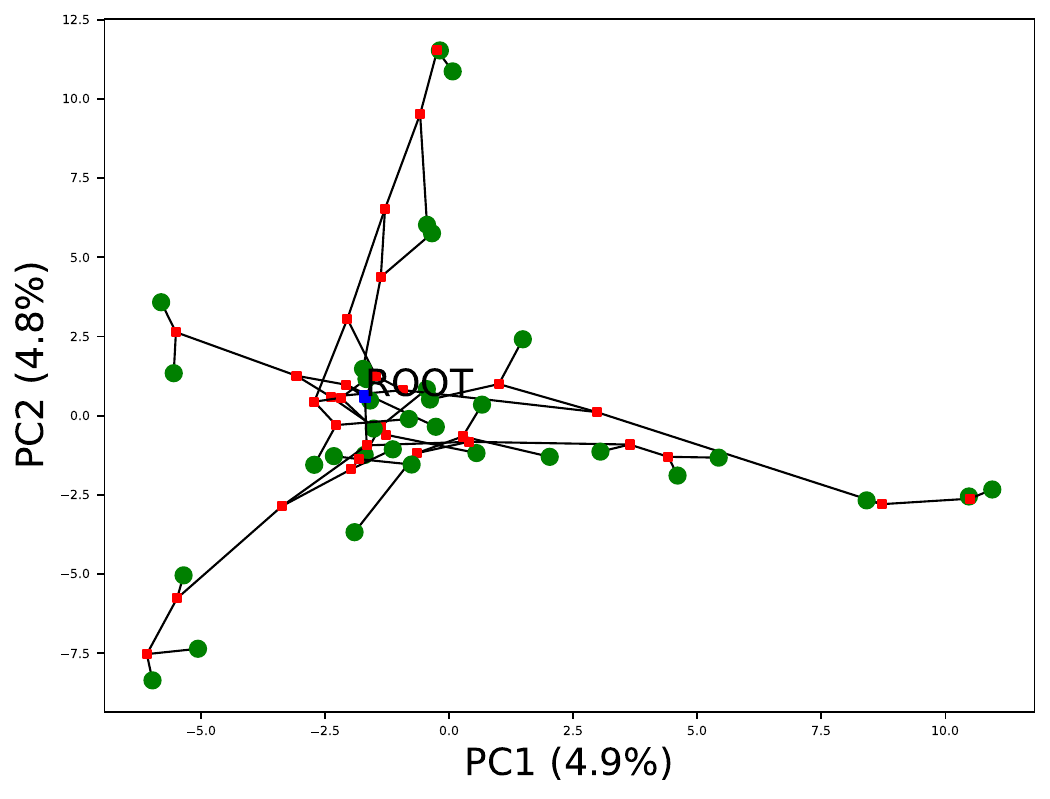}}\caption{PCA for the balanced time-tree of data simulation, with four borrowing scenarios.}\label{fig:d}\end{figure*}

We anticipate that ancestor-descendant transitions will predominantly follow a unidirectional pattern along the first PC axis. This expectation aligns with the tree model's implication of continuous diversification. Note, however, that this is not a rigid mathematical statement. To see why, let us consider three languages situated along a downward path in the tree. There are $2^3$ possible combinations for each of the hundreds or thousands of binary features. The two changeless patterns (\texttt{0} $\rightarrow$ \texttt{0} $\rightarrow$ \texttt{0} and \texttt{1} $\rightarrow$ \texttt{1} $\rightarrow$ \texttt{1}) can be ignored. The four single-change patterns (\texttt{0} $\rightarrow$ \texttt{0} $\rightarrow$ \texttt{1}, \texttt{0} $\rightarrow$ \texttt{1} $\rightarrow$ \texttt{1}, \texttt{1} $\rightarrow$ \texttt{0} $\rightarrow$ \texttt{0}, and \texttt{1} $\rightarrow$ \texttt{1} $\rightarrow$ \texttt{0}) together contribute to unidirectionality. Among the remaining two patterns, \texttt{0} $\rightarrow$ \texttt{1} $\rightarrow$ \texttt{0} is a perfectly valid transition and yet goes against unidirectionality. The last pattern, \texttt{1} $\rightarrow$ \texttt{0} $\rightarrow$ \texttt{1}, is a violation of the assumption, with horizontal transmission being the main contributing factor, although sporadic parallel innovations cannot be entirely dismissed.

Recall that PCA is a linear transformation, and $\boldsymbol{u}_1$ acts as a weight vector for mean-shifted feature sequences. If the evolutionary process is indeed tree-like, we can ignore the last pattern and anticipate the dominance of the four progressive patterns over the first regressive pattern. The loss of a feature is expected to be largely compensated by the gain of another feature because every language is expected to have at least one word for a basic vocabulary item. To conclude, a gross violation of the unidirectionality, which we call \textit{jogging}, can be seen as a deviation from the tree model.

Note that the absence of visible violations does not automatically imply the validity of the model for given data. Additionally, in the event that anomalies are detected, there is no feasible way to rescue the tree model. Therefore, the proposed method should primarily serve as a sanity check.

One obvious limitation of the proposed method is that it works on a single sample although Bayesian analysis conventionally draws conclusions by summarizing multiple samples. While applying PCA to multiple trees is possible, visualizing the outcome remains a challenge. If we focus on a specific clade, we can visualize a summary of multiple samples, as we see in Section~\ref{sec:g}.

The proposed method enables the verification of results from published papers. Note that slight modifications to the existing configuration files of Bayesian inference are required. This is necessary because, as mentioned in Section~\ref{sec:b}, the sampler does not track the states of unobserved languages by default. Technical details will be provided in Appendix~\ref{sec:j}.\section{Simulation Experiments}\subsection{Data Simulation}We evaluated the proposed method using synthetic data. To generate the data, we partly followed the procedure described by \citet{n}. We obtained the same skewed and balanced time-trees (Supplementary Figure~\ref{fig:k}) and used the software package TraitLab~\citep{ah} to simulate evolutionary processes along the branches of each time-tree, with or without borrowing of features between branches.

TraitLab implemented the stochastic Dollo model, which assumes that a feature can only be gained once in history and that once lost in a branch, it is never regained by descendants. This assumption is suitable for simulation of lexical items although it is considered too stringent when fitting real data~\citep{ai}.

TraitLab supported two borrowing scenarios for simulation. One was the global borrowing scenario, enabling borrowings among any contemporary languages, and the other is the local borrowing scenario which allowed borrowings only when the two languages shared a common ancestor within a specified time period. For each time-tree, we tested four scenarios: (1)~no borrowing, (2)~global borrowing, (3)~local borrowing with the 1,000-year limit, and (4)~local borrowing with the 3,000-year limit.\begin{figure*}[htbp]\centering\subfigure{\includegraphics[width=0.47\textwidth]{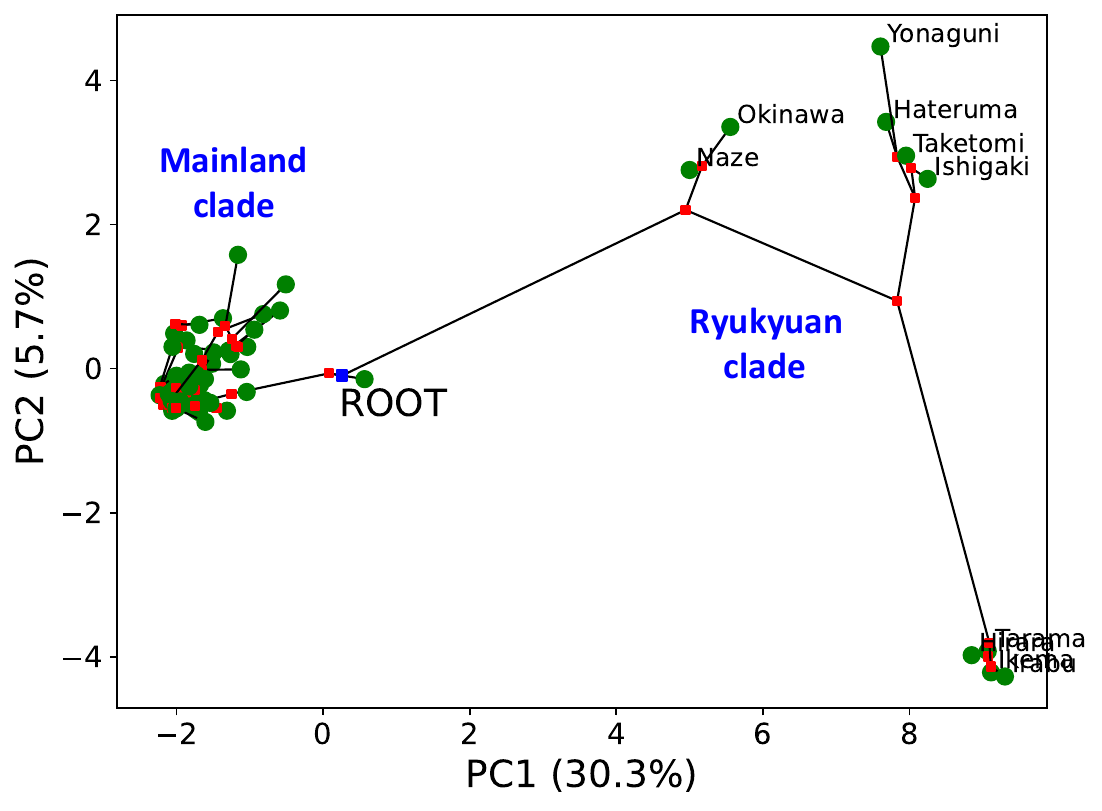}}\hfill\subfigure{\includegraphics[width=0.51\textwidth]{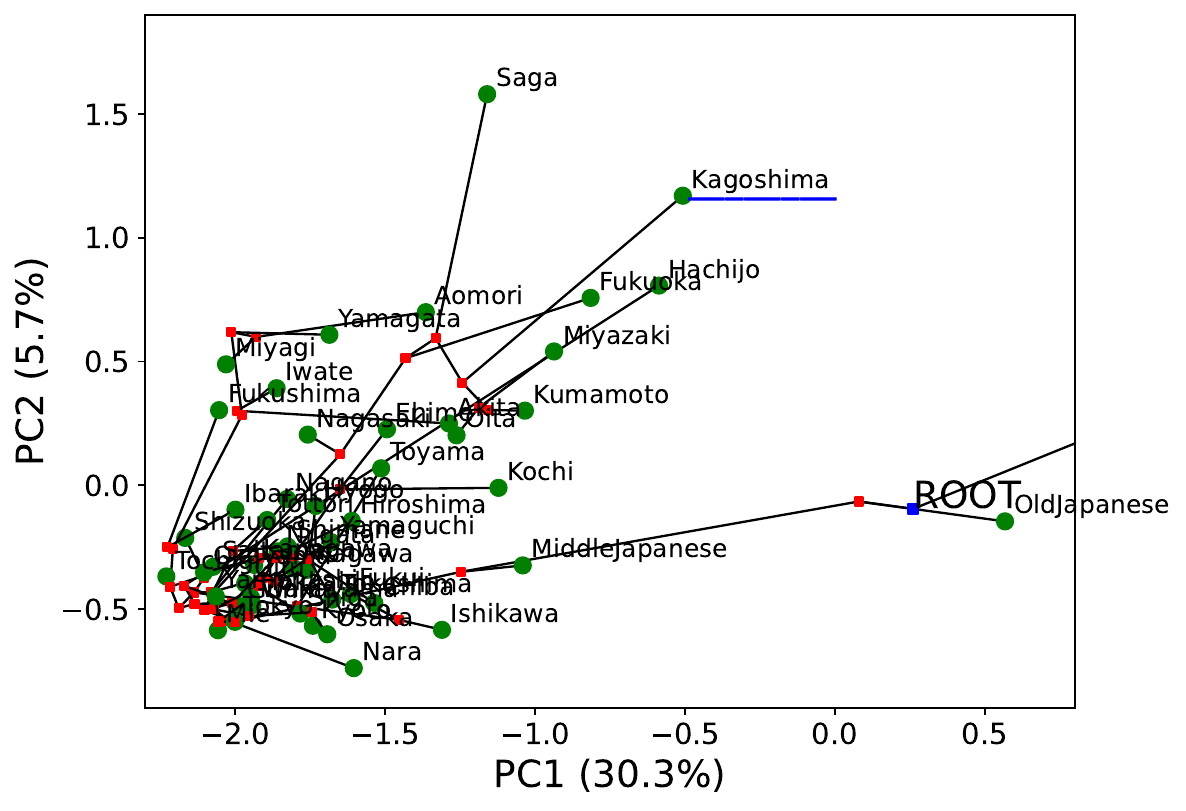}}\caption{PCA for a Japonic sample tree. Left: The entire tree. Right: Zoomed-in view of the mainland portion. Kagoshima (underlined) is the closest modern mainland dialect to Old Japanese along PC1.}\label{fig:e}\end{figure*}

Regarding hyperparameters, we configured the loss rate to be $0.2$ per 1,000 years and the mean number of traits (features) to be $200$. For borrowing scenarios, we set the borrowing rate at $2.241$, indicating that as many as 50\% of features were borrowed along an evolutionary path within a span of 1,000 years.\subsection{Phylolinguistic Reconstruction}We used the software package BEAST 2.7.5~\citep{y} to reconstruct the evolutionary process from observed languages. For simplicity, we used a Yule tree prior as the time-tree model, a binary continuous-time Markov chain model as the state transition model, and a strict clock as the rate model. Since age calibration was not conducted, we only estimated relative dates. We manually modified auto-generated configuration files to output the node states. We perform MCMC with a total of 10 million steps and applied PCA to the final sample.\subsection{Results}Figures~\ref{fig:c} and \ref{fig:d} show PCA projections of the tree samples. Our anticipation was validated by the synthetic data: In the absence of borrowing, the trees maintained near-perfect unidirectionality. In contrast, under the borrowing scenarios, all trees exhibited jogging.

The structural pattern observed under the no-borrowing scenario was better preserved in the balanced tree than in the skewed tree. This was likely due to the direct translation of high-level clades into the first two PCs.\section{Analyzing Real Data}\subsection{Japonic}We reviewed an analysis of the Japonic languages by \citet{aj}. Using basic vocabulary data from 59 Japonic dialects, they conducted a phylolinguistic tree reconstruction, with a primary emphasis on determining the root age. They contended that the estimated root age aligned with the putative agricultural population expansion of Japonic speakers.

A peculiarity of their approach was that they analyzed closely-related dialects that were usually considered to be primarily characterized by horizontal transmission~\citep{am}. To our knowledge, no one had applied the comparative method of historical-comparative linguistics to analyze their primary source, a dialect dictionary~\citeplanguageresource{an}.\footnote{A recent phylolinguistic reconstruction of Japonic languages~\citep{ao} is build on top of a careful manual selection of shared innovations, not a quantitative analysis of the entire lexical data.} Although \citet{aj} expressed some reservations about the non-tree-like nature of the data, they nonetheless persisted in utilizing the tree model.

Some effort was needed to replicate their analysis because no BEAST configuration file was published. We extracted binary-coded data from a supplementary PDF. We selected the model and hyperparameters based on the description of the paper although we replaced the relaxed clock model with a newer, more efficient one~\citep{ak}. Although several errors had been identified in the data~\citep{al}, we only corrected language names. Our MCC tree (Supplementary Figure~\ref{fig:l}) suggests that we replicated the original analysis to a large extent.\begin{figure*}[htbp]\centering\subfigure{\includegraphics[width=0.495\textwidth]{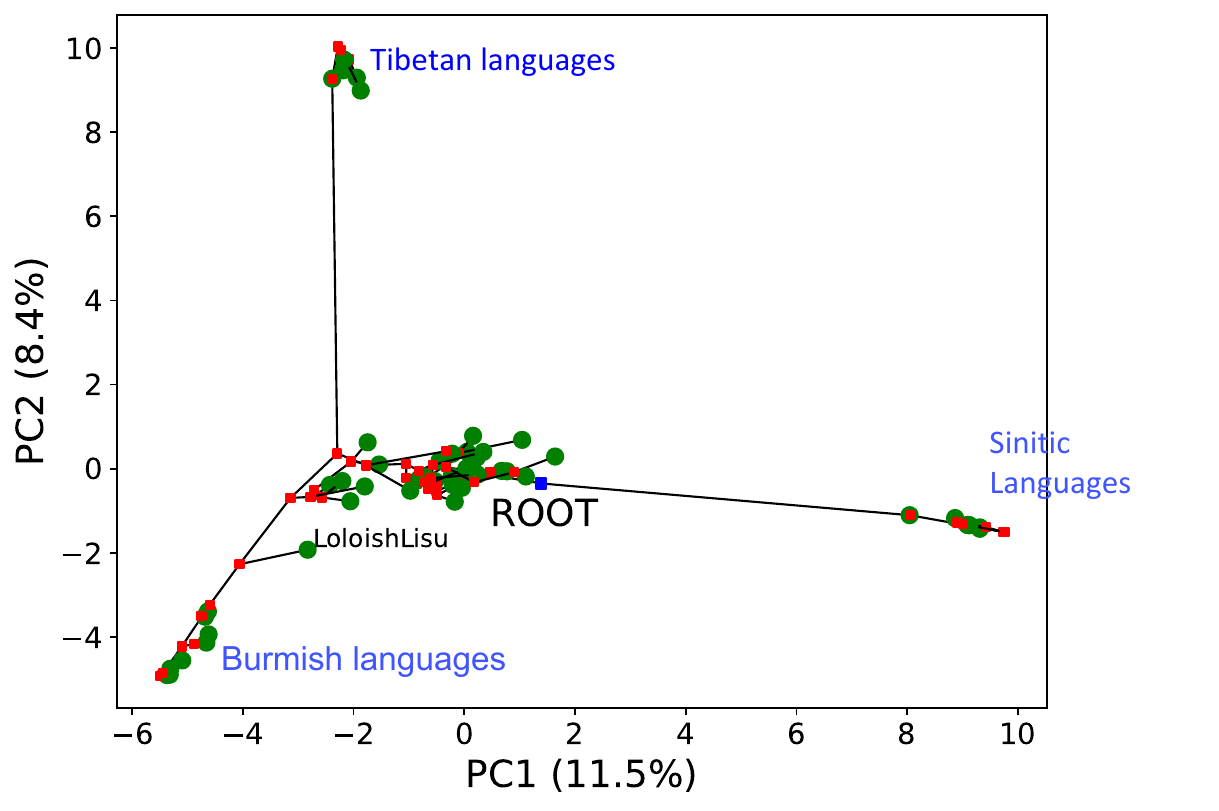}}\hfill\subfigure{\includegraphics[width=0.495\textwidth]{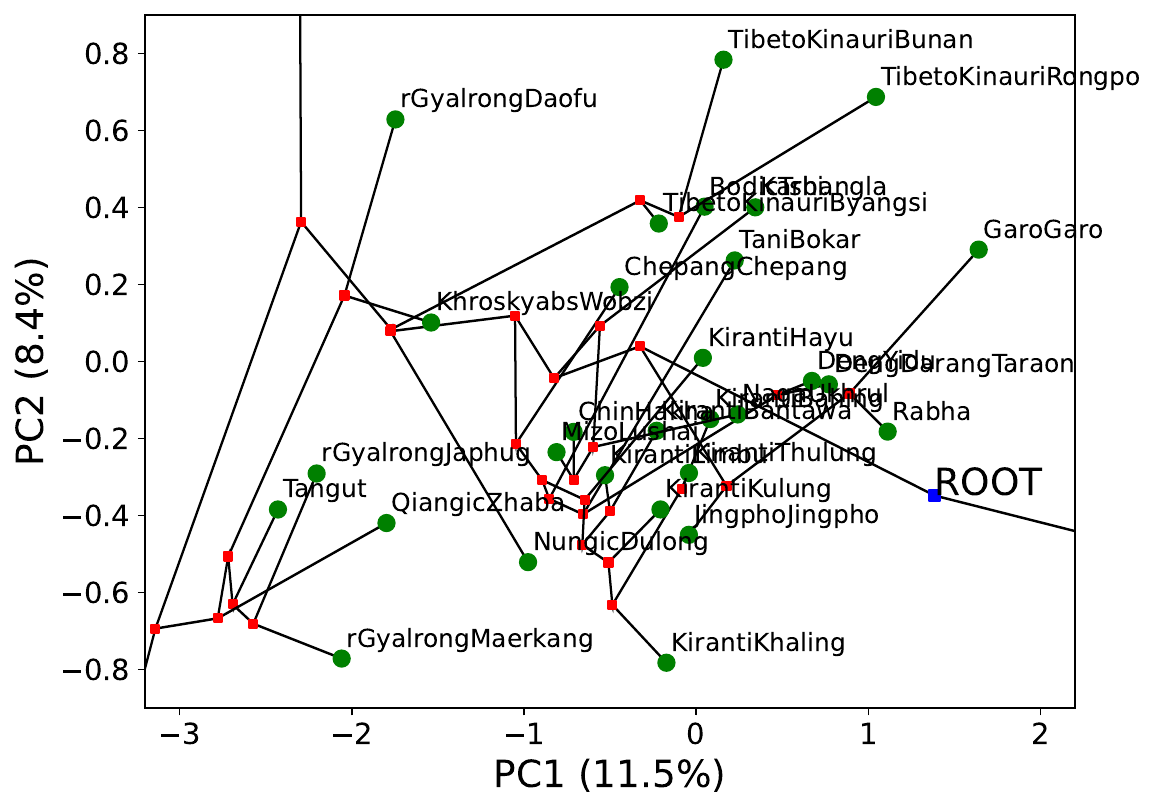}}\caption{PCA for a Sino-Tibetan sample tree. Left: The entire tree. Right: Zoomed-in view of the central portion.}\label{fig:f}\end{figure*}

Figure~\ref{fig:e} shows the PCA projection of the final tree sample. The first PC manifested a well-known division between the mainland and Ryukyuan, while also revealing considerable internal diversity within Ryukyuan. When examining the mainland, anomalies were evident. The extensive amount of jogging confirmed the inapplicability of the tree model to this dataset in an intuitive manner. The estimated root age is deemed unreliable because it was derived from the flawed trees.

Kagoshima, located at the southwestern tip of the mainland, exhibited the closest resemblance to Old Japanese along the first PC axis even though it ranked as the second least similar to Old Japanese among the mainland varieties if we switched to similarity based on binary sequences. A plausible explanation of this disparity is that the leftmost area of the figure was characterized by a multitude of overlapping diffusional patterns that covered vast areas but did not consistently reach their peripheries. In other words, Kagoshima underwent a relatively rapid change because it was less affected by dialect leveling, but the features it retained signaled archaism.\subsection{Sino-Tibetan}\label{sec:g}We turned our attention to \citet{ap}, who investigated Sino-Tibetan phylogenies. Also known as Trans-Himalayan, the Sino-Tibetan language family encompasses not just Chinese, Burmese, and Tibetan but also numerous smaller languages found in the mountainous regions of Asia.The high-level structure of Sino-Tibetan, including whether Sinitic represents a primary branch, remains poorly understood. Recent studies have also explored a potential connection between the emergence of Sino-Tibetan branches and the early phases of agriculture in northern China.

Sino-Tibetan is renowned for posing significant challenges in historical-comparative linguistics, with its complex contact history being a key factor~\citep{aq}. With the world's leading Sino-Tibetan specialists on their team, \citet{ap} carefully compiled a lexical database themselves and excluded from their analysis languages known for intense contact such as Bai. Our interest lies in assessing whether their sophisticated methodology effectively addressed the problem of horizontal transmission.\begin{figure}[t]\centering\includegraphics[width=0.495\textwidth]{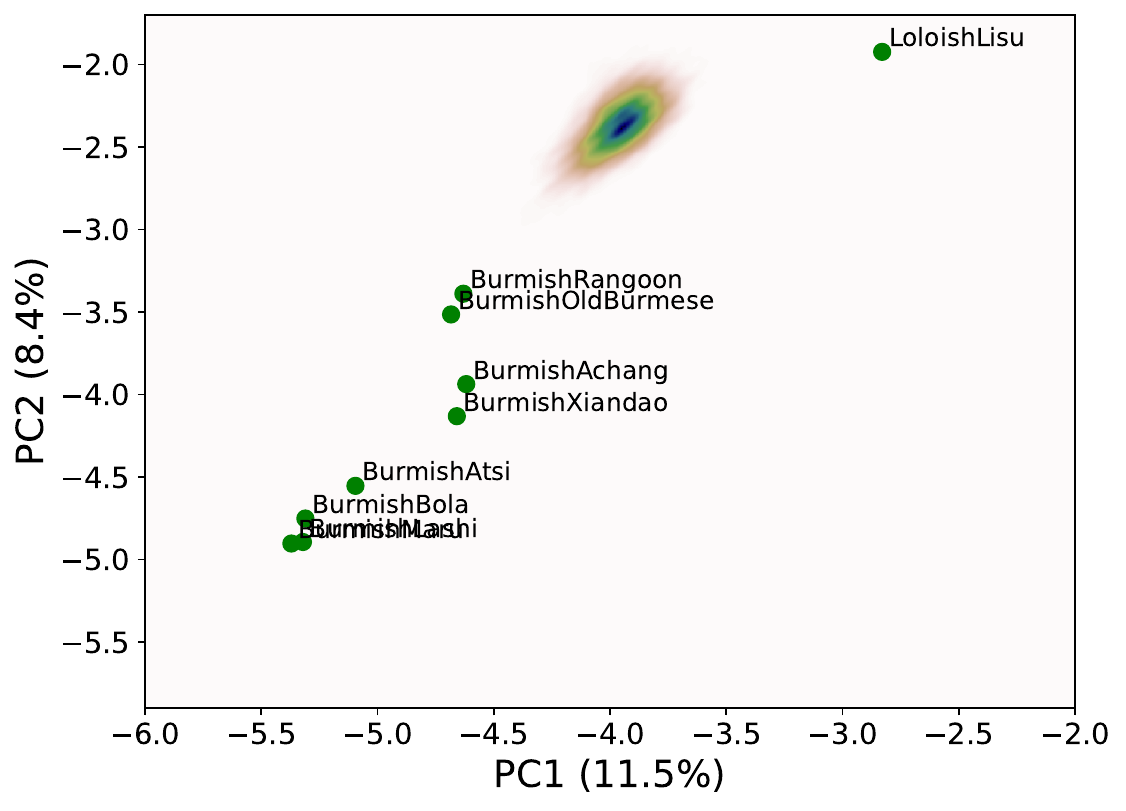}\caption{Kernel density estimation of the location of Proto-Lolo-Burmese.}\label{fig:h}\end{figure}\begin{figure*}[htbp]\centering\subfigure[Pseudo Dollo covarion model.]{\includegraphics[width=0.495\textwidth]{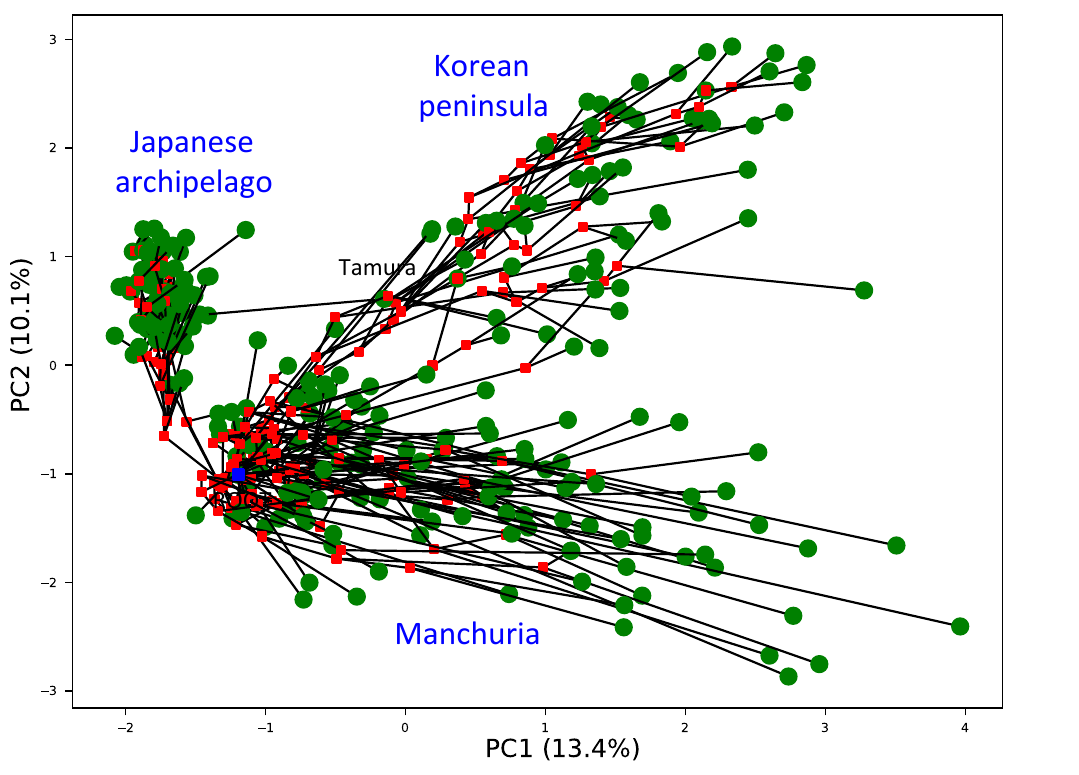}}\hfill\subfigure[Covarion model.]{\includegraphics[width=0.495\textwidth]{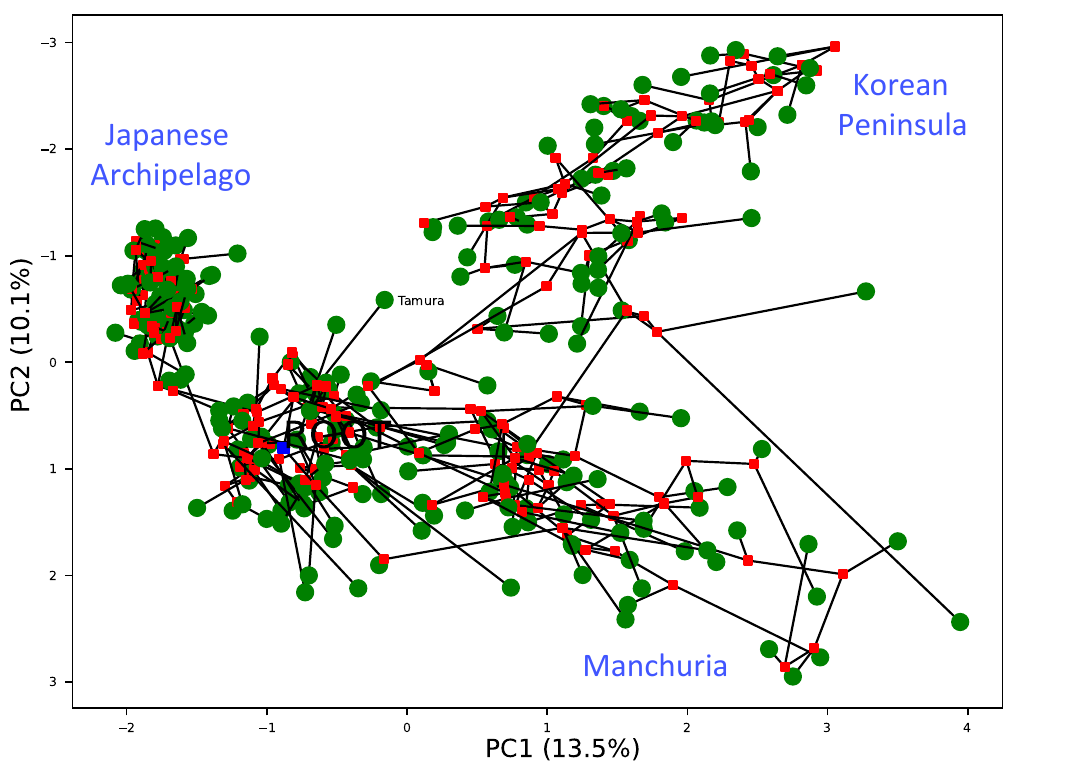}}\caption{PCA for the Northeast Asian archaeological sites. (a)~The model selected by \citet{ar} used the pseudo Dollo covarion model for state transition. (b)~The covarion model was used instead. The y-axis is inverted for the sake of facilitating comparison.}\label{fig:i}\end{figure*}

We used a BEAST configuration file\footnote{\texttt{sinotibetan-beast-covarion-relaxed\\-fbd.xml}} published as part of the supplementary materials and slightly modified it to output node states. Our MCC tree (Supplementary Figure~\ref{fig:m}) again indicates largely successful replication.

Figure~\ref{fig:f} shows our PCA projection of the final tree sample. Within this subspace, three distinct clusters emerge at the extremities, namely Sinitic, Tibetan, and Burmish, all of which had long writing traditions and were oversampled in the dataset. The remaining languages form a clustered group near the root and exhibit noticeable levels of jogging. According to the model, the evolutionary paths from Proto-Sino-Tibetan (the root) toward these languages follow a trajectory that includes Burmish-like intermediate nodes before moving back in the direction of the root. While Sinitic occupies the opposite end of the axis, the relative positions of these languages do not seem to correlate with their similarity to Sinitic.

We conducted a further analysis of the Loloish language of Lisu, which was located slightly outside the cluttered group. With the posterior probability of nearly 100\%, Lisu shared a direct common ancestor (Proto-Lolo-Burmese) with the Burmish languages. We collected the states of Proto-Lolo-Burmese from multiple samples and applied PCA projection. We then performed kernel density estimation to approximate the probability distribution of its location. The result is visualized in Figure~\ref{fig:h}. The phylolinguistic model demonstrated high confidence in determining the location of Proto-Lolo-Burmese, and thus in the presence of jogging in the evolutionary path to Lisu. Although we cannot conclude that the phylolinguistic reconstruction failed, the presence of anomalies necessitates further investigation.\subsection{Northeast Asian Archaeological Sites}Finally, we examined an analysis of archaeological sites of Northeast Asia by \citet{ar}, who advocated a version of the Altaic hypothesis under a new brand of Transeurasian. The highly controversial Altaic hypothesis posits a linguistic connection among Turkic, Mongolic, and Tungusic languages, and at times includes Koreanic and Japonic languages within this proposed single language family. It remains a minority view among historical linguists~\citep{as}.

A striking characteristic of \citet{ar} was their integration of archaeological, genetic, and linguistic evidence. However, all three types of evidence met biting criticism~\citep{at}. In this paper, we focused on the archaeological data because the apparent lack of tree-like signal was the focal point of criticism~\citep{at}.

We slightly edited a BEAST configuration file\footnote{\texttt{pdcov-ucln-bsp-tips.xml}} published as part of the supplementary materials. It contained 171 binary-coded (presence/absence) typological features of archaeology, such as pottery, horse, and wheat. We replaced coupled MCMC~\citep{au} with vanilla MCMC because the current implementation was incompatible with node state sampling. Comparing our MCC tree (Supplementary Figure~\ref{fig:n}) with the published result, we can observe that the two agreed on low-level groupings. There were disagreements on high-level groupings, but they can be explained by their extremely low posterior probabilities. Even if the tree model was applicable, the phylolinguistic model was highly uncertain about the high-level structure of the data.

The PCA projection of the final sample is shown in Figure~\ref{fig:i}(a). The first PC featured a distinction between Japan (left) and the rest of Northeast Asia (right). Overall, the projected tree revealed a pattern of continual diversification. A notable exception was Tamura, a site on Japan, which was buried in the Asian continent in the subspace despite being clearly descended from a Japanese parent. This can be interpreted as hybridization, a violation of the tree model. The MCC tree alone shows no sign of such a deviation.

The scarcity of jogging raises suspicion, as the data was perceived as markedly non-tree-like~\citep{at}. We argue that this stemmed from inappropriate model selection. The model selected by \citet{ar} used a pseudo Dollo model~\citep{ai} for state transition. This model loosely adheres to the Dollo principle, which suggests that a feature can be gained only once in a tree but lost multiple times. Because a na\"{i}ve implementation of this principle is highly sensitive to borrowings, the pseudo Dollo model permits multiple gains of a feature in a tree while it still restricts languages from reacquiring a feature that their ancestor had lost. \citet{ar} combined the pseudo Dollo model with the covarion model, which is widely used to capture fast and slow phases of evolution~\citep{av}.

The pseudo Dollo covarion model yields the complete absence of the \texttt{1} $\rightarrow$ \texttt{0} $\rightarrow$ \texttt{1} pattern, which strongly promotes unidirectionality. For comparison, we applied the PCA projection to the simple covarion model, based on the configuration file included in their supplementary materials.\footnote{\texttt{cov-strict-bsp.xml}} As expected, this model choice resulted in a substantial quantity of jogging~(Figure~\ref{fig:i}(b)).

While the the arbitrariness of meaning-symbol connection provides a rational basis for applying the Dollo principle to cognates, it is entirely plausible that a typological feature could potentially be reacquired. Although \citet{ar} justified their model choice based on its superior fit to the data, our analysis suggests that the apparent lack of jogging was an artifact of the inappropriate model selection.\section{Conclusions}In this paper, we have introduced a method for projecting a tree sample using principal component analysis in order to identify anomalies in Bayesian phylolinguistic reconstruction. A departure from the tree model can be observed as a deviation along the first principal component axis, which we refer to as jogging. The proposed method is strikingly simple and can be applied to a wide range of published data. Our primary focus is on binary-coded lexical data, as their meaning-symbol connection inherently enforces a unidirectional pattern under the tree model. Conducting a more comprehensive analysis of our approach's effectiveness on different data types would be a valuable avenue for further research.\section{Acknowledgments}We express our gratitude to Simon J. Greenhill for generously providing the code and data necessary for replicating the findings in \citet{n}. This work received partial support from JSPS KAKENHI Grant Numbers 21K12029 and 18KK0012.\section{Limitations}We investigated a critical assumption inherent in Bayesian phylolinguistic models, which is frequently violated in real-world scenarios. Our method aims to visualize deviations from the tree model, yet it is important to note that the absence of apparent violations does not guarantee the model's validity for the given data. Furthermore, in cases where anomalies are detected, there is no feasible way to rescue the tree model.

Principal component analysis is a parameter-free technique that operates under minimal assumptions. Nonetheless, it can be susceptible to bias when confronted with an overrepresentation of one or more clades within the dataset, potentially resulting in a skewed data representation.\nocite{*}\section{Bibliographical References}\bibliographystyle{lrec-coling2024-natbib}\bibliography{a}\section{Language Resource References}\bibliographystylelanguageresource{lrec-coling2024-natbib}\bibliographylanguageresource{b}\appendix\renewcommand\thefigure{A.\arabic{figure}}\setcounter{figure}{0}\section{Implementation Notes}\label{sec:j}To sample node states in the software package BEAST, we usually need to modify the existing configuration file. Specifically, we need to replace \texttt{TreeWithMetaDataLogger} with \texttt{AncestralSequenceLogger}. The ``logger'' does not just write logs but samples node states. The node states are output as node annotations in the NEXUS format. The logger requires the \texttt{tag} attribute specifying the key for NEXUS node annotations, the \texttt{data} attribute specifying the alignment data, the \texttt{siteModel} attribute specifying the site model, and the \texttt{branchRateModel} attribute specifying the branch rate model.

\texttt{AncestralSequenceLogger} is old and is included in the \texttt{beast-classic} package. It might not be compatible with newer modules.

Several recent studies define multiple site models to account for varying rates associated with basic vocabulary items. In such instances, a straightforward solution is to define a logger for each site model. Consequently, multiple copies of the same tree are generated, each providing distinct information about the node states. A postprocessing step is necessary to merge them into a single tree with complete node states.

Node state sampling can also be accomplished using the commonly utilized software BayesTraits~\citep{aw}, which effectively models state transitions for a given tree or set of tree samples. While theoretically feasible to apply our method to analyses based on BayesTraits, it is important to acknowledge that our approach necessitates multiple features, whereas BayesTraits is often employed to analyze a singular feature.\begin{figure*}[htbp]\centering\subfigure{\includegraphics[width=0.48\textwidth]{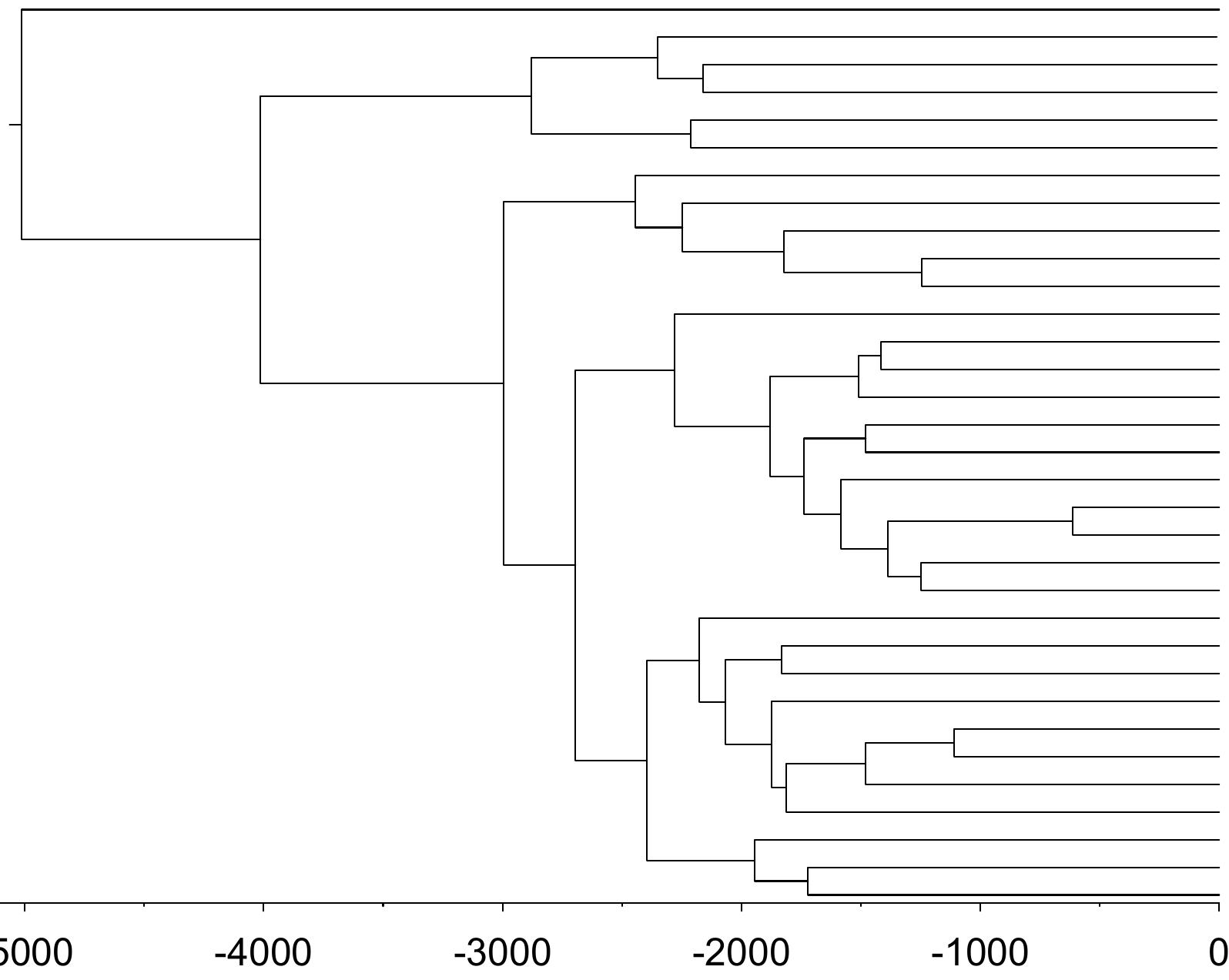}}\hfill\subfigure{\includegraphics[width=0.48\textwidth]{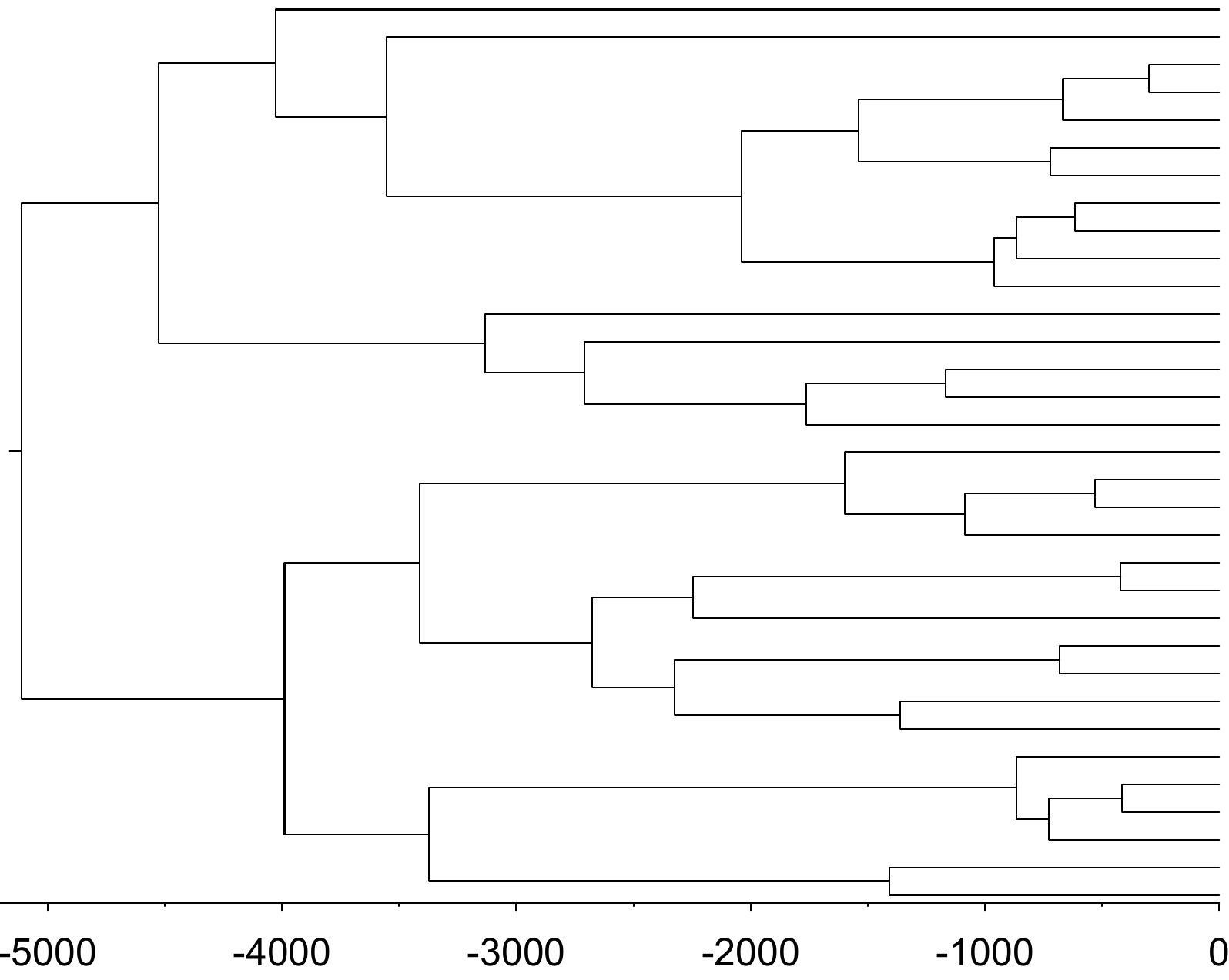}}\caption{Two time-trees used for data simulation by \citet{n}. One is skewed while the other is balanced. The horizontal axis represents the passage of time, measured in years.}\phantomsection\label{fig:k}\end{figure*}\begin{figure*}[htbp]\centering\includegraphics[width=0.98\textwidth]{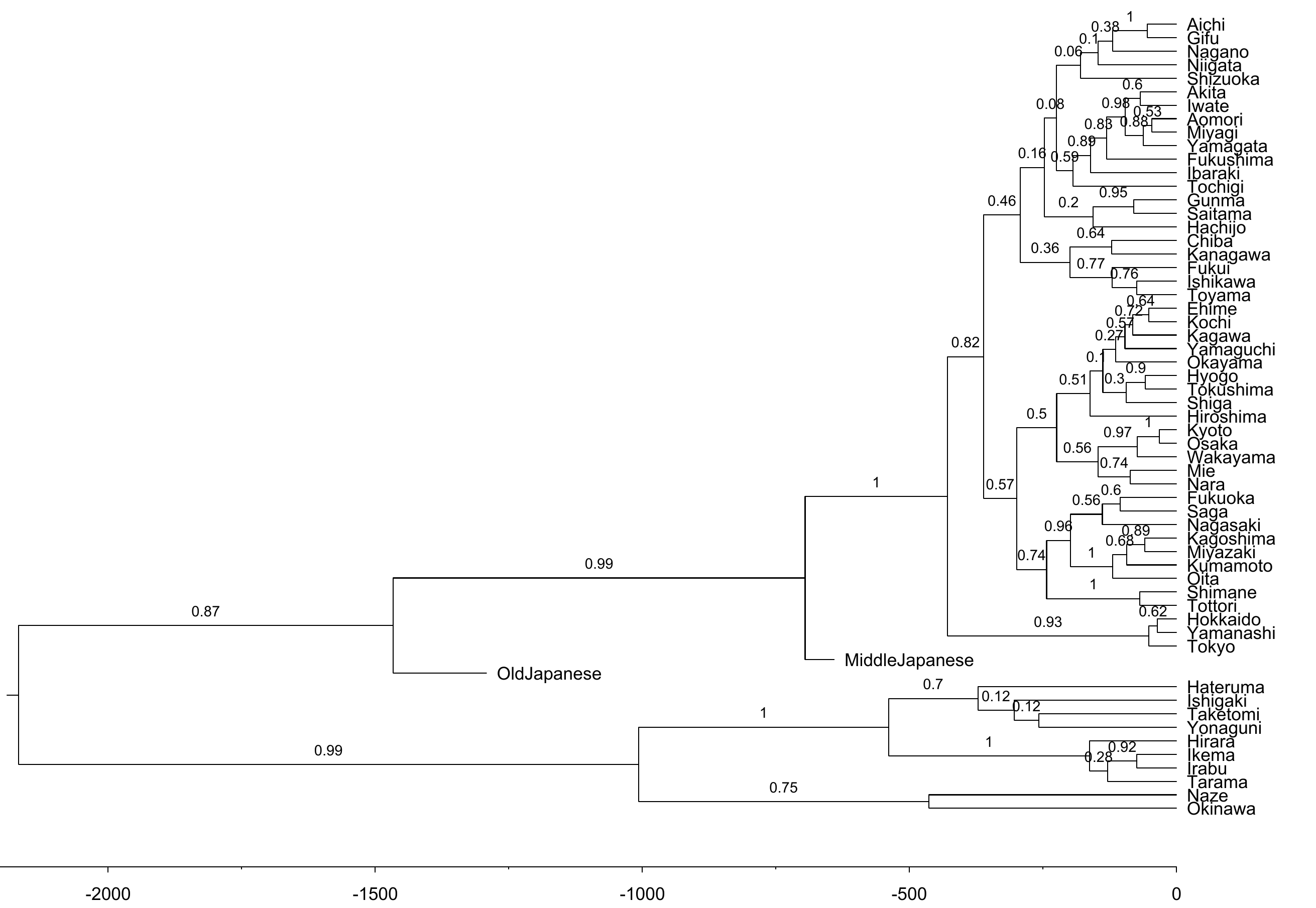}\caption{The maximum clade credibility tree of the Japonic languages. A number positioned above a branch indicates the posterior probability of the corresponding clade.}\phantomsection\label{fig:l}\end{figure*}\begin{figure*}[htbp]\centering\includegraphics[width=0.98\textwidth]{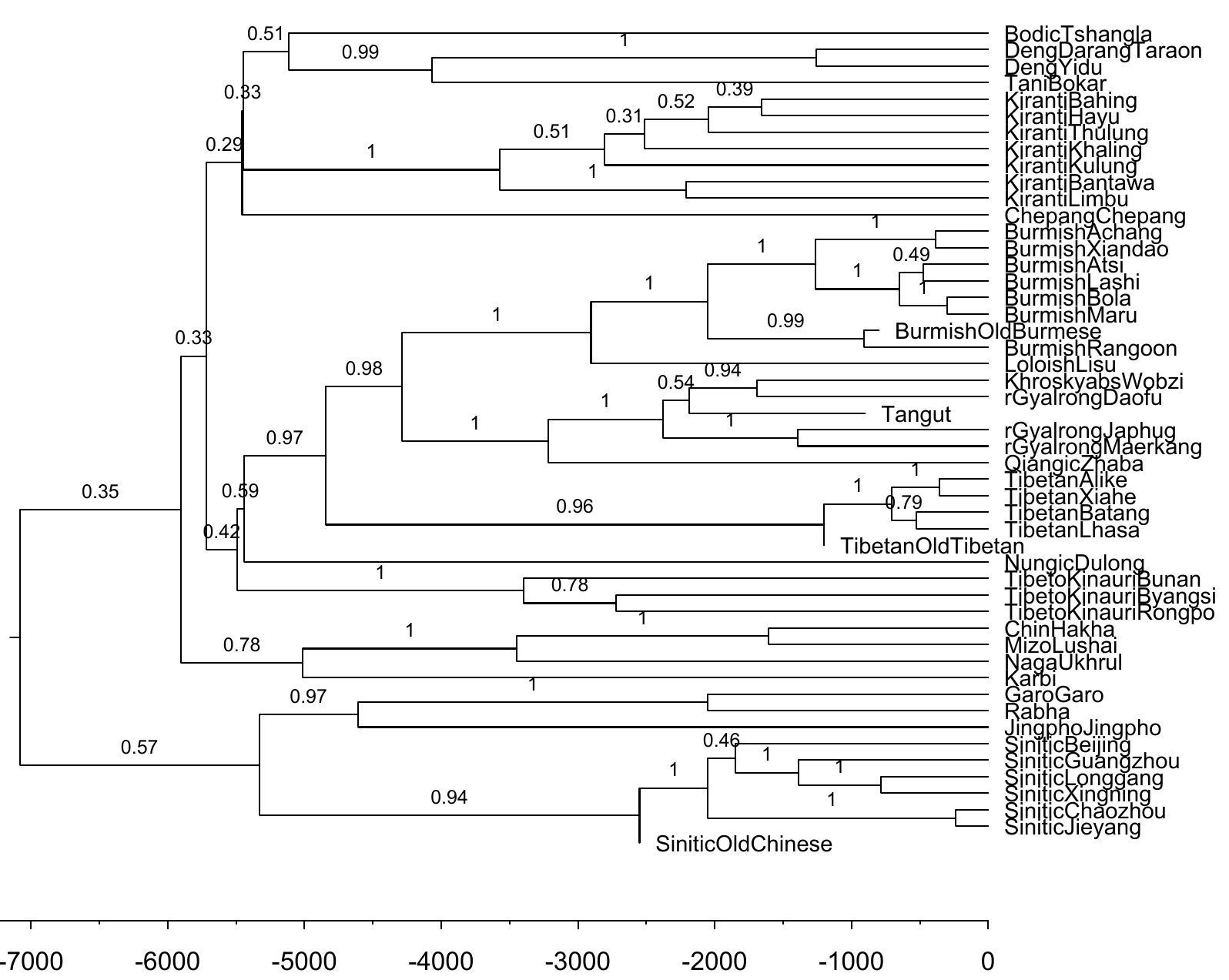}\caption{The maximum clade credibility tree of the Sino-Tibetan languages. A number positioned above a branch indicates the posterior probability of the corresponding clade.}\phantomsection\label{fig:m}\end{figure*}\begin{figure*}[htbp]\centering\rotatebox{90}{\includegraphics[width=1.25\textwidth]{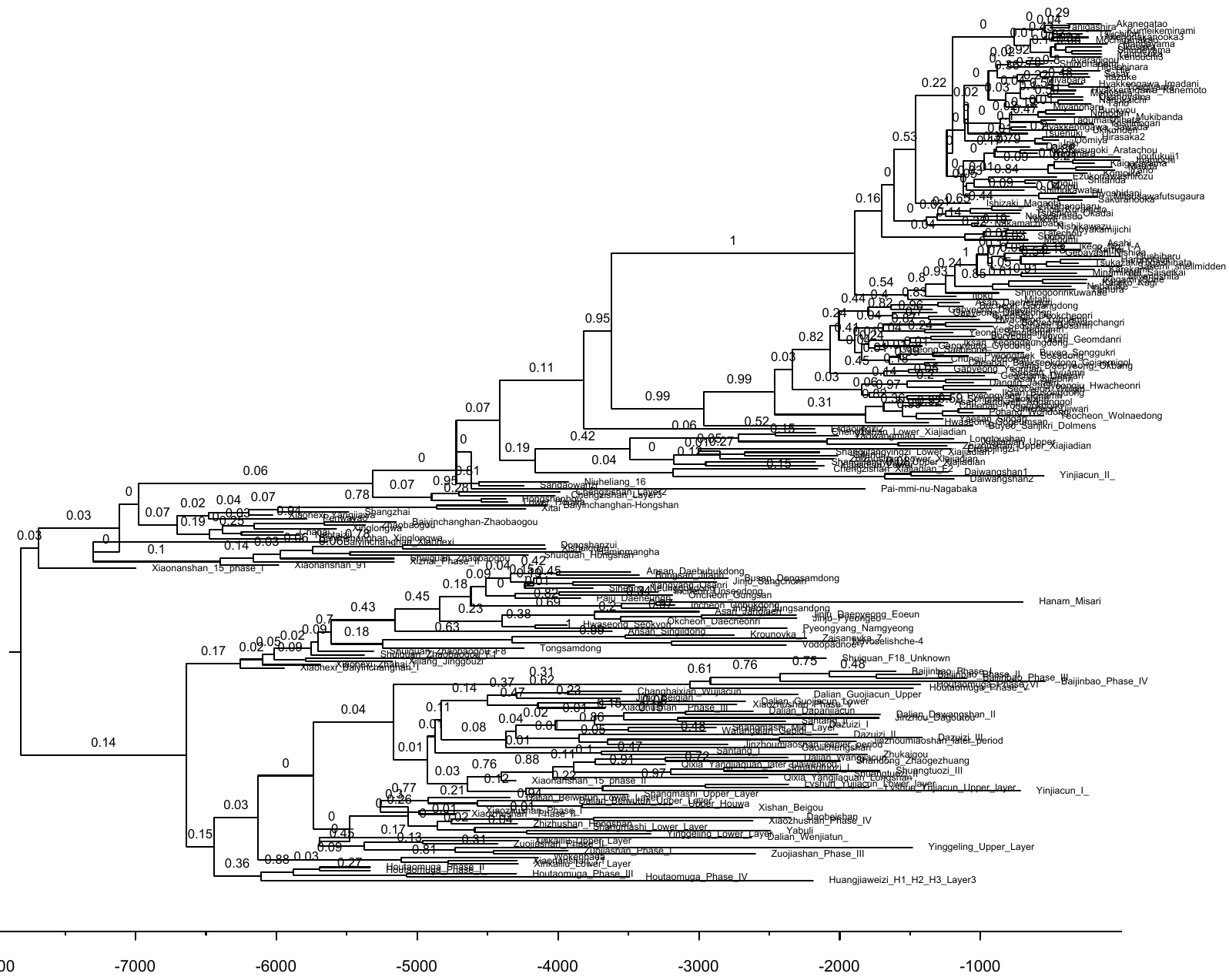}}\caption{The maximum clade credibility tree of the Northeast Asian archaeological sites. A number positioned above a branch indicates the posterior probability of the corresponding clade.}\phantomsection\label{fig:n}\end{figure*}\end{document}